\documentclass[10pt,twocolumn,letterpaper]{article}

\usepackage{iccv}
\usepackage{times}
\usepackage{epsfig}
\usepackage{graphicx}
\usepackage{amsmath}
\usepackage{amssymb}
\usepackage{float}
\usepackage{epstopdf}
\usepackage{caption}
\usepackage{subcaption}
\usepackage{color}
\usepackage[titletoc,title]{appendix}

\DeclareMathOperator*{\argmax}{arg\,max}

\usepackage{hyperref}

\iccvfinalcopy

\newcommand{\Section}[1]{\vspace{-0pt}\section{#1}\vspace{-0pt}}
\newcommand{\SubSection}[1]{\vspace{-0pt}\subsection{#1}\vspace{-0pt}}

\newcommand{\Caption}[1]{\vspace{-0pt}\caption{\em #1}\vspace{-0pt}}

\begin{document}

\title{Patch-based Convolutional Neural Network for Whole Slide Tissue Image Classification}

\author{
Le Hou$^1$, Dimitris Samaras$^1$, Tahsin M. Kurc$^{2}$,
       Yi Gao$^{2,1,3}$, James E. Davis$^4$, and Joel H. Saltz$^{2,1,4,5}$
       \\
       $^1$Department of Computer Science, Stony Brook University, NY, USA\\
       \{lehhou,samaras\}@cs.stonybrook.edu\\
       $^2$Department of Biomedical Informatics, Stony Brook University, NY, USA\\
       \{tahsin.kurc,joel.saltz\}@stonybrook.edu\\
       $^3$Department of Applied Mathematics and Statistics, NY, USA\\
       $^4$Department of Pathology, Stony Brook Hospital, NY, USA\\
       \{yi.gao,james.davis\}@stonybrookmedicine.edu\\
       $^5$Cancer Center, Stony Brook Hospital, NY, USA\\
}

\maketitle

\begin{abstract}
\vspace{-0.07cm}
Convolutional Neural Networks (CNN) are state-of-the-art models for many image classification tasks. However, to recognize cancer subtypes automatically, training a CNN on gigapixel resolution Whole Slide Tissue Images (WSI) is currently computationally impossible. The differentiation of cancer subtypes is based on cellular-level visual features observed on image patch scale. Therefore, we argue that in this situation, training a patch-level classifier on image patches will perform better than or similar to an image-level classifier. The challenge becomes how to intelligently combine patch-level classification results and model the fact that not all patches will be discriminative. We propose to train a decision fusion model to aggregate patch-level predictions given by patch-level CNNs, which to the best of our knowledge has not been shown before. Furthermore, we formulate a novel Expectation-Maximization (EM) based method that automatically locates discriminative patches robustly by utilizing the spatial relationships of patches. We apply our method to the classification of glioma and non-small-cell lung carcinoma cases into subtypes. The classification accuracy of our method is similar to the inter-observer agreement between pathologists. Although it is impossible to train CNNs on WSIs, we experimentally demonstrate using a comparable non-cancer dataset of smaller images that a patch-based CNN can outperform an image-based CNN.
\end{abstract}

\Section{Introduction}

\begin{figure}[ht]
\begin{center}
   \includegraphics[width=0.85\linewidth]{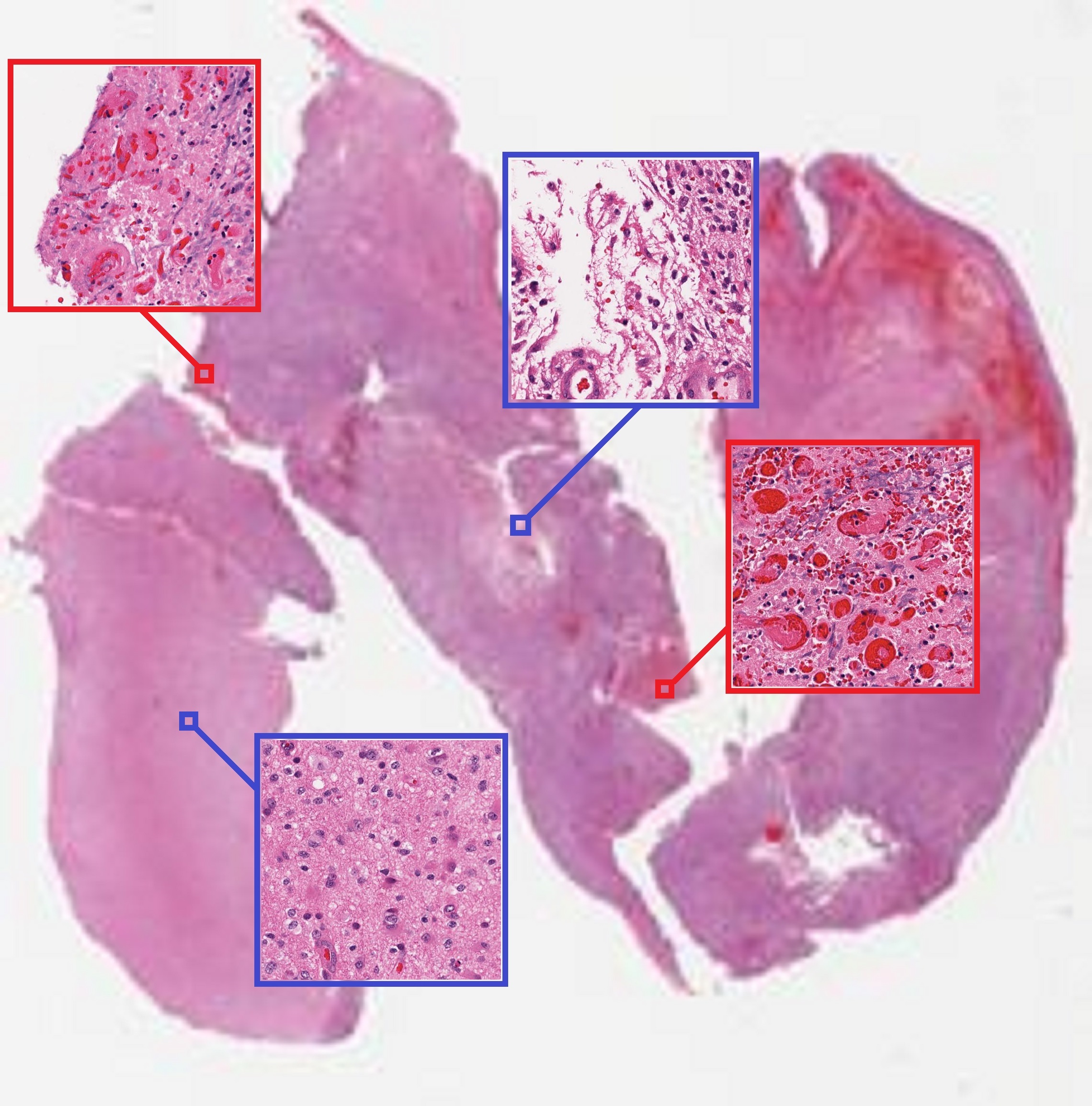}
\end{center}
   \Caption{A gigapixel Whole Slide Tissue Image of a grade IV tumor (best viewed in color). Visual features that determine the subtype and grade of a WSI are visible in high resolution details. In this case, patches framed in red are discriminative since they show typical visual features of grade IV tumor. Patches framed in blue are non-discriminative because they only contain visual features from lower grade tumors. Notice that discriminative patches are dispersed throughout the image at multiple locations.}
\label{fig:wsi}
\end{figure}

Convolutional Neural Networks (CNNs) are currently the state-of-the-art image classifiers~\cite{lecun1998gradient,krizhevsky2012imagenet,bengio2013representation,he2015delving}. However, due to high computational cost, CNNs cannot be applied to very high resolution images, such as gigapixel Whole Slide Tissue Images (WSI). Classification of cancer WSIs into grades and subtypes is critical to the study of disease onset and progression and the development of targeted therapies, because the effects of cancer can be observed in WSIs at the cellular and sub-cellular levels (Fig.~\ref{fig:wsi}). Applying CNN directly for WSI classification has several drawbacks. First, extensive image downsampling is required by which most of the discriminative details could be lost. Second, it is possible that a CNN might only learn from one of the multiple discriminative patterns in an image, resulting in data inefficiency. Discriminative information is encoded in high resolution image patches. Therefore, one solution is to train a CNN on high resolution image patches and predict the label of a WSI based on patch-level predictions.

The ground truth labels of individual patches are unknown, as only the image-level ground truth label is given. This complicates the classification problem. Because tumors may have a mixture of structures and texture properties, patch-level labels are not necessarily consistent with the image-level label. More importantly, when aggregating patch-level labels to an image-level label, simple decision fusion methods such as voting and max-pooling are not robust and do not match the decision process followed by pathologists. For example, a mixed subtype of cancer such as oligoastrocytoma, might have distinct regions of other cancer subtypes. Therefore, neither voting nor max-pooling could predict the correct WSI-level label since the patch-level predictions do not match the WSI-level label.

\begin{figure}[ht]
\begin{center}
   \fbox{\includegraphics[trim=40mm 33mm 37mm 26mm, clip, width=0.98\linewidth]{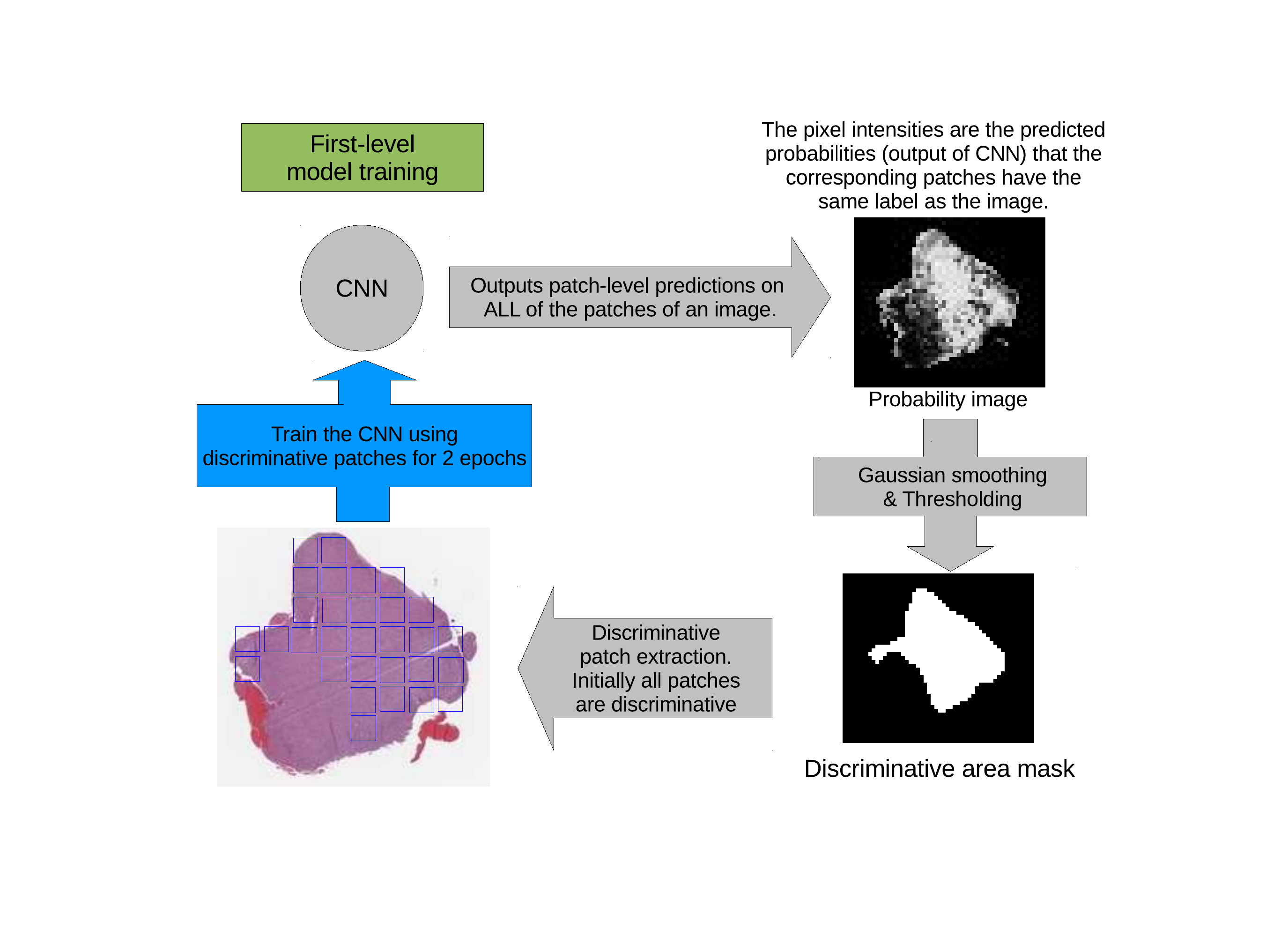}}
   \fbox{\includegraphics[trim=33mm 26mm 41mm 21mm, clip, width=0.98\linewidth]{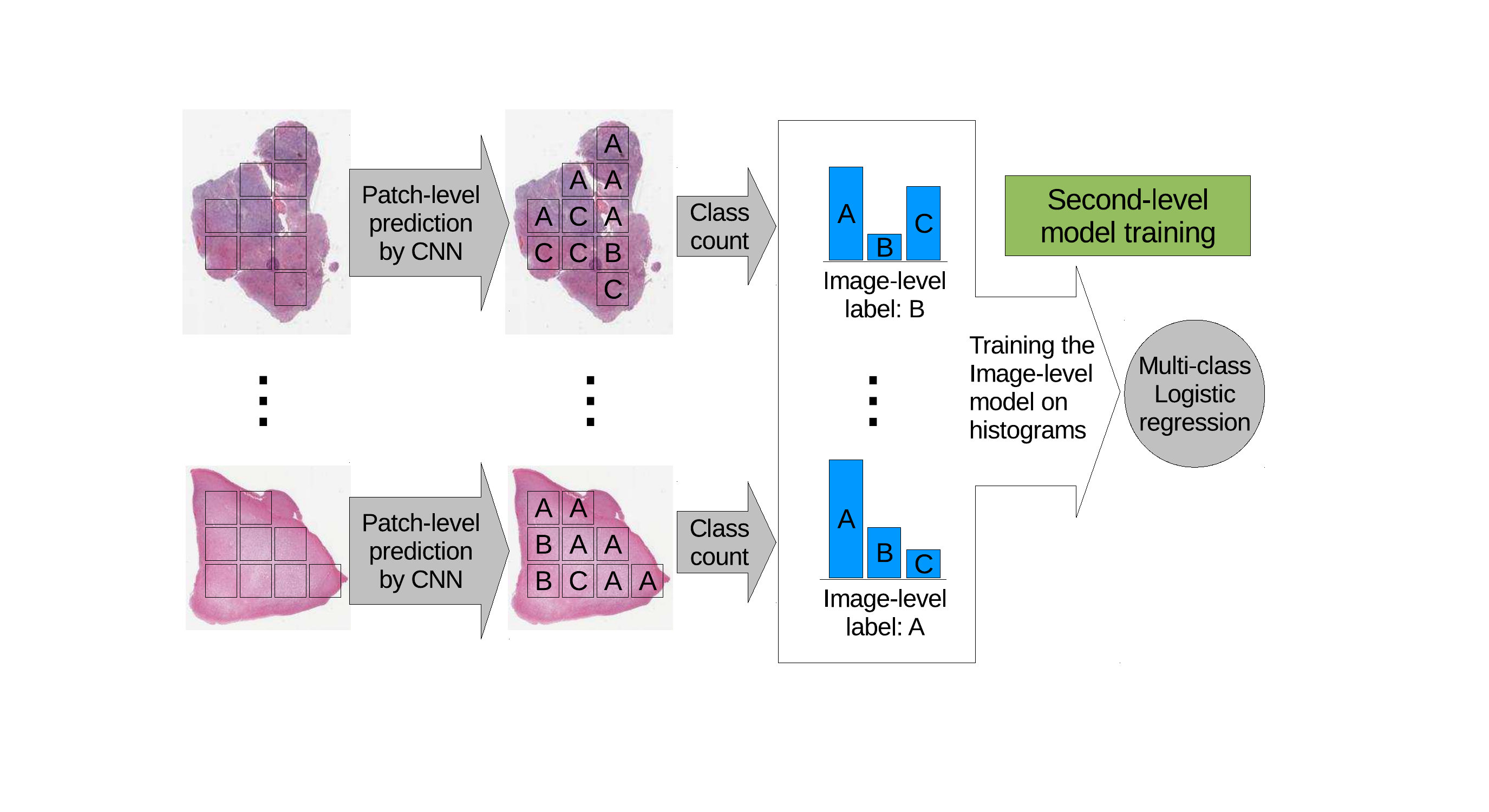}}
\end{center}
   \Caption{An overview of our workflow (best viewed in color). Top: A CNN is trained on patches. An EM-based method iteratively identifies non-discriminative patches and eliminates them from the CNN training set. Bottom: An image-level decision fusion model is trained on histograms of patch-level predictions, to predict the image-level label.}
\label{fig:workflow}
\end{figure}

We propose using a patch-level CNN and training a decision fusion model as a two-level model, shown in Fig.~\ref{fig:workflow}. The first-level (patch-level) model is an Expectation Maximization (EM) based method combined with CNN that outputs patch-level predictions. In particular, we assume that there is a hidden variable associated with each patch extracted from an image that indicates whether the patch is discriminative or not. Here, ``a discriminative patch'' means that the true hidden label of the patch is the same as the true label of the image. Initially, we consider all patches to be discriminative. We train a CNN model that outputs the cancer type probability of each input patch. We apply spatial smoothing to the resulting probability map and select only patches with higher probability values as discriminative patches. We iterate this process using the new set of discriminative patches in an EM fashion until convergence. In the second-level (image-level), histograms of patch-level predictions are input into an image-level multiclass logistic regression or Support Vector Machine (SVM)~\cite{chang2011libsvm} model that predicts the image-level labels.

Pathology image classification and segmentation is an active field of research. Most WSI classification methods focus on classifying or extracting features on patches~\cite{cruz2014automatic,mousavi2015automated,xu2015deep,zhou2014classification,chang2014stacked,altunbay2010color,vu2015dfdl,cirecsan2013mitosis,xu2015deep}. In~\cite{xu2015deep}  a pretrained CNN model  extracts features on patches which are then aggregated  for WSI classification. As shown by our experiments, the heterogeneity of some cancer subtypes cannot be captured by those generic CNN features. Patch-level supervised classifiers can learn the heterogeneity of cancer subtypes, if a lot of patch labels are provided~\cite{cruz2014automatic,mousavi2015automated}. However, acquiring such labels in large scale has prohibitive cost, due to the need for highly  specialized annotators. As digitization of tissue samples becomes increasingly commonplace, one can envision large scale datasets, that could not be annotated at patch scale. Utilizing unlabeled patches has led to Multiple Instance Learning (MIL) based WSI classification~\cite{cosatto2013automated,xu2014deep,xu2014weakly}.

In the MIL paradigm~\cite{dietterich1997solving,maron1998framework,amores2013multiple}, unlabeled instances belong to labeled bags of instances. The goal is to predict the label of a new bag and/or the label of each instance. The Standard Multi-Instance (SMI) assumption ~\cite{dietterich1997solving} states that for a binary classification problem, a bag is positive iff there exists at least one positive instance in the bag. The probability of a bag being positive equals to the maximum positive prediction over all of its instances~\cite{andrews2002support,zhang2005multiple,kim2010gaussian}. Combining MIL with Neural Networks (NN)~\cite{ramon2000multi,zhou2002neural,li2012efficient,chen2013multi}, the SMI assumption is modeled by max-pooling. Following this formulation, the Back Propagation for Multi-Instance Problems (BP-MIP)~\cite{ramon2000multi,zhou2002neural} performs back propagation along the instance with the maximum response if the bag is positive. This is inefficient because only one instance per bag is trained in one training iteration on the whole bag.

MIL-based CNNs have been applied to object recognition~\cite{oquab2014weakly} and semantic segmentation~\cite{pathak2014fully} in image analysis -- the image is the bag and image-windows are the instances~\cite{nguyen2009weakly}.  
These methods also follow the SMI assumption. The training error is only propagated through the object-containing window which is also assumed to be the window that has the maximum prediction confidence. This is not robust because one significantly misclassified window might be considered as the object-containing window. Additionally, in WSIs, there might be multiple windows that contain discriminative information. Recent semantic image segmentation approaches~\cite{chen2014semantic,pinheiro2014weakly,papandreou2015weakly} 
smooth the output probability (feature) maps of the CNNs. In this way, they can identify relevant windows more robustly.

To predict the image-level label, max-pooling (SMI) and voting (average-pooling) were applied in~\cite{nguyen2009weakly,lecun1998gradient,cruz2014automatic}. However, it has been shown that in many applications, learning decision fusion models can significantly improve  performance compared to voting~\cite{poria2015deep,seff20142d,hoai2015improving,tabib2013decision,karpathy2014large,Simonyan14c}. Furthermore, such a learned decision fusion model is based on the Count-based Multiple Instance (CMI) assumption which is the most general MIL assumption~\cite{weidmann2003two}.

Our main contributions in this paper are:
\begin{enumerate}
\item To the best of our knowledge, we are the first to combine patch-level CNNs with supervised decision fusion. Aggregating patch-level CNN predictions for WSI classification significantly outperforms patch-level CNNs with max-pooling or voting.
\item We propose a new EM-based model that identifies discriminative patches in high resolution images automatically for patch-level CNN training, utilizing the spatial relationship between patches.
\item Our model achieves multiple state-of-the-art results classifying WSIs to cancer subtypes on the TCGA dataset. Our results are similar or close to inter-observer agreement between pathologists. Larger classification improvements are observed in the harder-to-classify cases.
\item We provide experimental evidence that combining multiple patch-level classifiers might actually be advantageous compared to whole image classification.
\end{enumerate}

The rest of this paper is organized as follows. Sec.~\ref{sec:our_method1} describes the framework of the EM-based MIL algorithm. Sec.~\ref{sec:our_method2} discusses the identification of discriminative patches. Sec.~\ref{sec:our_method3} explains the image-level model that predicts the image-level label by aggregating patch-level predictions. Sec.~\ref{sec:experiments} shows experimental results. The paper concludes in Sec.~\ref{sec:conclusions}. App.~\ref{app:app} lists the cancer subtypes in our experiments.

\Section{EM-based method with CNN}
\label{sec:our_method1}

An overview of our EM-based method can be found in Fig.~\ref{fig:workflow}. We model a high resolution image as a bag and patches extracted from it as instances. We have a ground truth label for the whole image but not for the individual patches. 
We model whether an instance is discriminative or not as a hidden binary variable.

We denote $X=\{X_1, X_2, \dots, X_N\}$ as the dataset containing $N$ bags. Each bag $X_i=\{X_{i,1}, X_{i,2}, \dots, X_{i,N_i}\}$ consists of $N_i$ instances, where $X_{i,j} = \langle x_{i,j} ,\text{ } y_{i} \rangle$ is the $j$-th instance and its associated label in the $i$-th bag. Assuming the bags are independent and identically distributed (i.i.d.), the $X$ and the hidden variables $H$ are generated by the following generative model:
\begin{equation}
\begin{split}
P(X, H) &= \prod\limits_{i=1}^{N} \Big( P(X_{i,1}, \dots, X_{i,N_i} \mid H_i) P(H_i) \Big) \text{,}
\end{split}
\label{equ:generative_model}
\end{equation}
where the hidden variable $H=\{H_1, H_2, \dots, H_N\}$, $H_i=\{H_{i,1}, H_{i,2}, \dots, H_{i,N_i}\}$ and $H_{i,j}$ is the hidden variable that indicates whether instance $x_{i,j}$ is discriminative  for  label $y_{i}$ of bag $X_i$. We further assume that all $X_{i,j}$ depends on $H_{i,j}$ only and are independent with each other given $H_{i,j}$. Thus
\begin{equation}
P(X, H) = \prod\limits_{i=1}^{N} \prod\limits_{j=1}^{N_i} \Big( P(X_{i,j} \mid H_{i,j}) P(H_i) \Big) \text{.}
\label{equ:conditional_independence}
\end{equation}

We maximize the data likelihood $P(X)$ using EM.
\begin{enumerate}
\setlength\itemsep{0.04em}
\item At the initial E step, we set $H_{i,j}=1$ for all $i, j$. This means that all instances are considered discriminative.
\item M step: We update the model parameter $\theta$ to maximize the data likelihood
\begin{equation}
\begin{split}
\theta \leftarrow & \argmax\limits_{\theta} P(X \mid H; \theta) \\
= & \argmax\limits_{\theta} \prod\limits_{x_{i,j} \in D} P(x_{i,j}, y_i \mid \theta) \\
& \times \prod\limits_{x_{p,q} \not \in D} P(x_{p,q}, y_q \mid \theta)\text{,}
\end{split}
\label{equ:m_step1}
\end{equation}
where $D$ is the discriminative patches set. Assuming a uniform generative model for all non-discriminative instances, the optimization in Eq.~\ref{equ:m_step1}  simplifies to:
\begin{equation}
\begin{split}
& \argmax\limits_{\theta} \prod\limits_{x_{i,j} \in D} P(x_{i,j}, y_i \mid \theta) \\
= & \argmax\limits_{\theta} \prod\limits_{x_{i,j} \in D} P(y_i \mid x_{i,j}; \theta) P(x_{i,j} \mid \theta) \text{.}
\end{split}
\label{equ:m_step2}
\end{equation}
Additionally we assume an uniform distribution over $x_{i,j}$. Thus Eq.~\ref{equ:m_step2} describes a discriminative model (in this paper we use a CNN).
\item E step: We estimate the hidden variables $H$. In particular, $H_{i,j}=1$ if and only if $P(H_{i,j}\mid X)$ is above a certain threshold. In the case of image classification, given the $i$-th image, $P(H_{i,j}\mid X)$ is obtained by applying Gaussian smoothing on $P(y_i\mid x_{i,j};\theta)$ (Detailed in Sec~\ref{sec:our_method2}). This smoothing step utilizes the spatial relationship of $P(y_i\mid x_{i,j}; \theta)$ in the image. We then iterate back to the M step till convergence.
\end{enumerate}

Many MIL algorithms can be interpreted through this formulation. Based on the SMI assumption, the instance with the maximum $P(H_{i,j}\mid X)$ is the discriminative instance for the positive bag, as in the EM Diverse Density (EM-DD)~\cite{zhang2001dd} and the BP-MIP~\cite{ramon2000multi,zhou2002neural} algorithms.

\Section{Discriminative patch selection}
\label{sec:our_method2}
Patches $x_{i,j}$ that have $P(H_{i,j}\mid X)$ larger than a threshold $T_{i,j}$ are considered discriminative and are selected to continue training the CNN. We present in this section the estimation of $P(H\mid X)$ and the choice of the threshold. 

It is reasonable to assume that $P(H_{i,j}\mid X)$ is correlated with $P(y_{i} \mid x_{i,j}; \theta)$, \textit{i.e}, patches with lower $P(y_{i} \mid x_{i,j}; \theta)$ tend to have lower probability $x_{i,j}$ to be discriminative. However, a hard-to-classify patch, or a patch close to the decision boundary may have low $P(y_{i} \mid x_{i,j}; \theta)$ as well. These patches are informative and should not be rejected. Therefore, to obtain a more robust $P(H_{i,j}\mid X)$, we apply the following two steps: First, we train two CNNs on two different scales in parallel. $P(y_{i} \mid x_{i,j}; \theta)$ is the averaged prediction of the two CNNs. Second, we simply denoise the probability map $P(y_{i} \mid x_{i,j}; \theta)$ of each image with a Gaussian kernel to compute $P(H_{i,j}\mid X)$.

Choosing a thresholding scheme carefully yields significantly better performance than a simpler thresholding scheme~\cite{papandreou2015weakly}. We obtain the threshold $T_{i,j}$ for $P(H_{i,j} \mid X)$ as follows: We note $S_i$ as the set of $P(H_{i,j}\mid X)$ values for all $x_{i,j}$ of the $i$-th image and $E_c$ as the set of $P(H_{i,j}\mid X)$ values for all $x_{i,j}$ of the $c$-th class. We introduce the image-level threshold $H_i$ as the $P_1$-th percentile of $S_i$ and the class-level threshold $R_i$ as the $P_2$-th percentile of $E_c$, where $P_1$ and $P_2$ are predefined. The threshold $T_{i,j}$ is defined as the minimum value between $H_i$ and $R_i$. There are two advantages of our method. First, by using the image-level threshold, there are at least $1-P_1$ percent of patches that are considered discriminative for each image. Second, by using the class-level threshold, the thresholds can be easily adapted to classes with different prior probabilities.

\Section{Image-level decision fusion model}
\label{sec:our_method3}

We combine the patch-level classifiers of Sec.~\ref{sec:our_method2} to predict the image-level label. We input all patch-level predictions into a multi-class logistic regression or SVM that outputs the image-level label. This decision level fusion method~\cite{kokar2001data} is  more robust than max-pooling~\cite{seff20142d}. Moreover, this method can be thought of as a Count-based Multiple Instance (CMI) learning method with two-level learning~\cite{weidmann2003two} which is a more general MIL assumption~\cite{foulds2010review} than the Standard Multiple Instance (SMI) assumption.

There are three reasons for combining multiple instances: First, on difficult datasets, we do not want to assign an image-level prediction simply based on a single patch-level prediction (as is the case of the SMI assumption~\cite{dietterich1997solving}). Second, even though certain patches are not discriminative individually, their joint appearance might be discriminative. For example, a WSI of the ``mixed'' glioma, Oligoastrocytoma (see App.~\ref{app:app}) should be recognized when two single glioma subtypes (Oligodendroglioma and Astrocytoma) are jointly present on the slide possibly on non-overlapping regions. Third, because the patch-level model is never perfect and probably biased, an image-level decision fusion model may learn to correct the bias of patch-level decisions.

In particular, the class histogram of the patch-level predictions is the input to a linear multi-class logistic regression model~\cite{bishop2006pattern} or an SVM with Radial Basis Function (RBF) kernel~\cite{chang2011libsvm}. To generate the histogram, we simply sum up all of the class probabilities given by the patch-level CNN. Moreover, we concatenate histograms from four CNNs models: CNNs trained at two patch scales for two different numbers of iterations. We found in practice that concatenating multiple histograms is robust.

\Section{Experiments}
\label{sec:experiments}
We evaluate our method on two Whole Slide Tissue Images (WSI) classification problems: classification of glioma and Non-Small-Cell Lung Carcinoma (NSCLC) cases into glioma and NSCLC subtypes. Glioma is a type of brain cancer that rises from glial cells. It is the most common malignant brain tumor and the leading cause of cancer-related deaths in people under age 20~\cite{brain_tumor_facts}. NSCLC is the most common lung cancer, which is the leading cause of cancer-related deaths overall~\cite{nsclc}. Classifying glioma and NSCLC into their respective subtypes and grades is crucial to the study of disease onset and progression in order to provide targeted therapies. The dataset of WSIs used in the experiments is composed from the public Cancer Genome Atlas (TCGA) dataset~\cite{tcga}. It contains detailed clinical information and the Hematoxylin and Eosin (H\&E) stained images of various cancers. The typical resolution of a WSI in this dataset is 100K by 50K pixels. 
In the rest of this section, we first describe the algorithm we tested then show the evaluation results on the glioma and NSCLC classification tasks.

\begin{figure*}[t]
\begin{center}
   \begin{subfigure}[b]{0.15\textwidth}
   	\includegraphics[width=\textwidth]{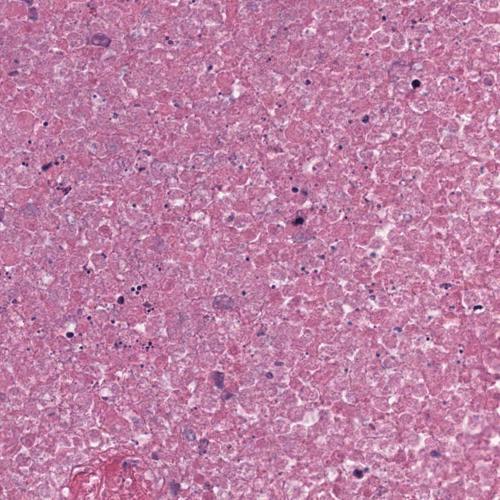}
   \end{subfigure}
   \begin{subfigure}[b]{0.15\textwidth}
   	\includegraphics[width=\textwidth]{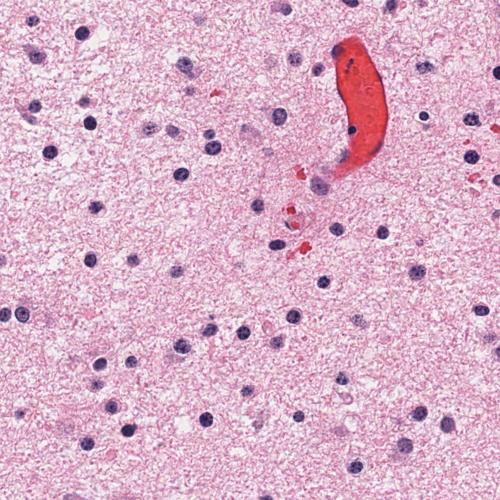}
   \end{subfigure}
   \begin{subfigure}[b]{0.15\textwidth}
   	\includegraphics[width=\textwidth]{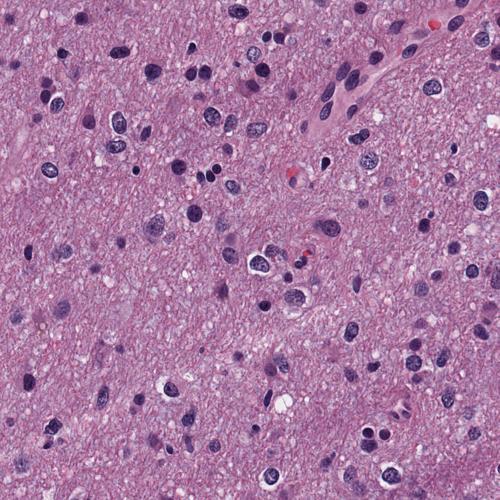}
   \end{subfigure}
   \begin{subfigure}[b]{0.15\textwidth}
   	\includegraphics[width=\textwidth]{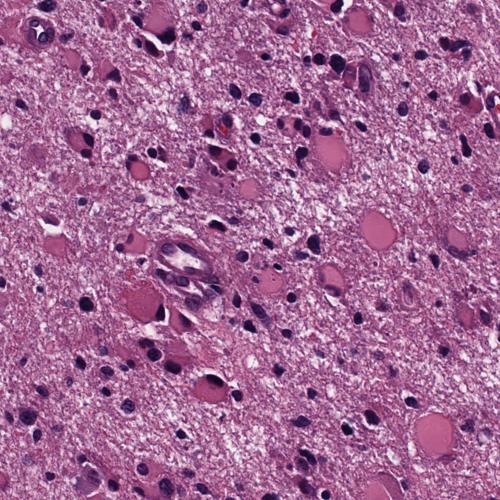}
   \end{subfigure}
   \begin{subfigure}[b]{0.15\textwidth}
   	\includegraphics[width=\textwidth]{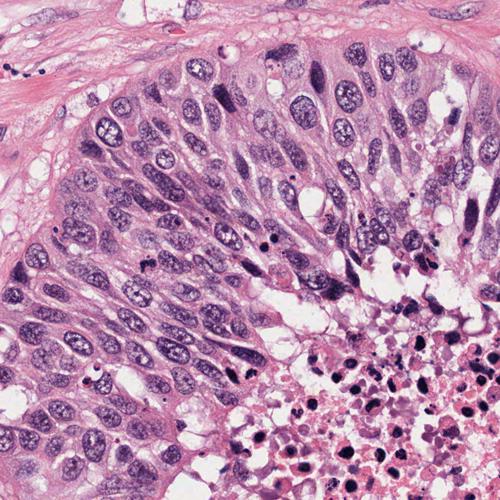}
   \end{subfigure}
   \begin{subfigure}[b]{0.15\textwidth}
   	\includegraphics[width=\textwidth]{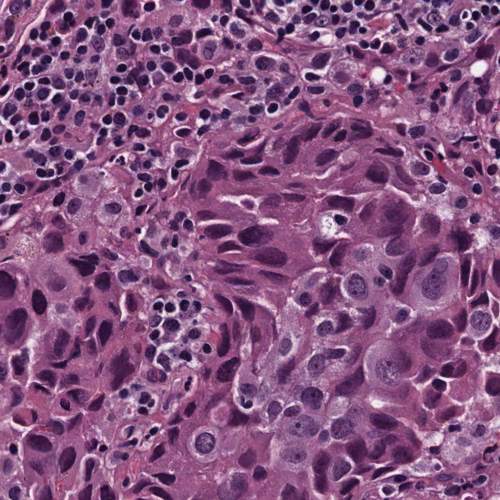}
   \end{subfigure}
   \vspace{0.1cm}
   
   \begin{subfigure}[b]{0.15\textwidth}
   	\includegraphics[width=\textwidth]{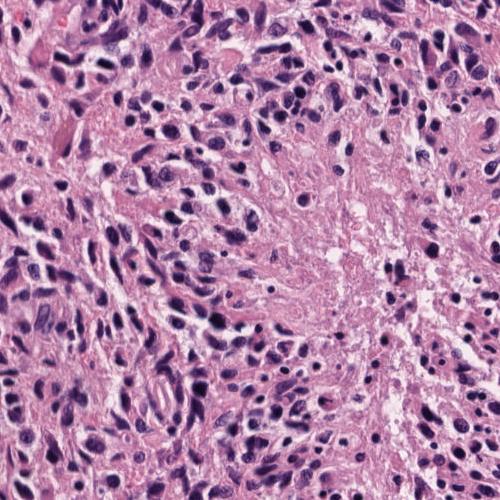}
    \vspace{-0.3cm}
   	\Caption{GBM}
   \end{subfigure}
   \begin{subfigure}[b]{0.15\textwidth}
   	\includegraphics[width=\textwidth]{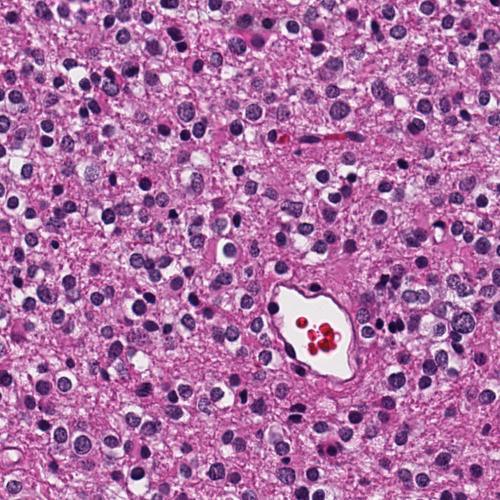}
    \vspace{-0.3cm}
   	\Caption{OD}
   \end{subfigure}
   \begin{subfigure}[b]{0.15\textwidth}
   	\includegraphics[width=\textwidth]{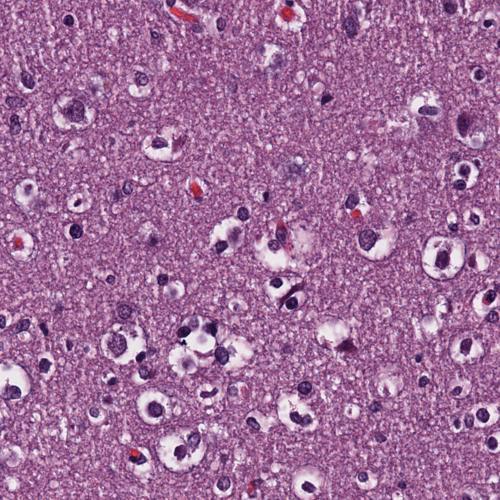}
    \vspace{-0.3cm}
   	\Caption{OA}
   \end{subfigure}
   \begin{subfigure}[b]{0.15\textwidth}
   	\includegraphics[width=\textwidth]{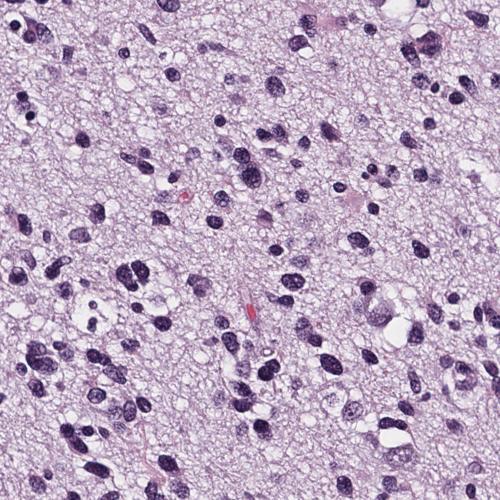}
    \vspace{-0.3cm}
   	\Caption{DA}
   \end{subfigure}
   \begin{subfigure}[b]{0.15\textwidth}
   	\includegraphics[width=\textwidth]{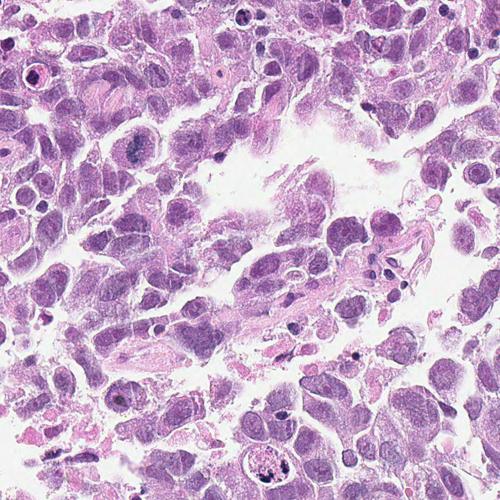}
    \vspace{-0.3cm}
   	\Caption{SCC}
   \end{subfigure}
   \begin{subfigure}[b]{0.15\textwidth}
   	\includegraphics[width=\textwidth]{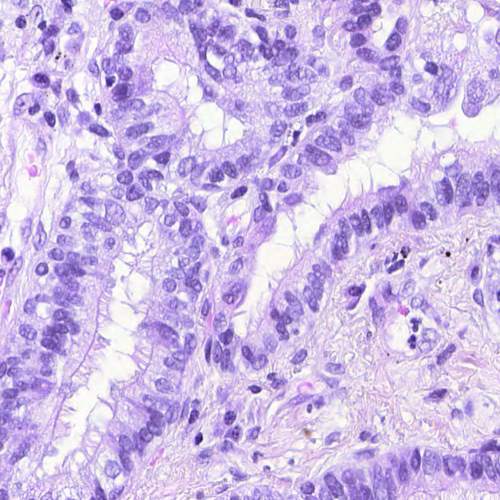}
    \vspace{-0.3cm}
   	\Caption{ADC}
   \end{subfigure}
\end{center}
\vspace{-0.5cm}
   \Caption{Some 20X sample patches of gliomas and Non-Small-Cell Lung Carcinoma (NSCLC) from the TCGA dataset. Two patches in each column belong to the same subtype of cancer. Notice  the large intra-class heterogeneity.}
\label{fig:patch_samples}
\end{figure*}
\SubSection{Patch extraction and segmentation}
To train the CNN model, patches of size 500 by 500 are extracted from WSIs. To capture structures at multiple scales, we extract patches from 20X (0.5 microns per pixel) and 5X (2.0 microns per pixel) objective magnifications. Patches that contain less than 30\% tissue sections or have too much blood are discarded. Around 1000 valid patches per image per scale are extracted. In most cases the patches are non-overlapping given the resolution of a WSI. Fig.~\ref{fig:patch_samples} shows sample patches.

To prevent the CNN from severe overfitting, we perform three kinds of data augmentation in every iteration. First, a random 400 by 400 sub-patch is selected from each 500 by 500 patch. Second, the sub-patch is randomly rotated and mirrored. Third, the amount of Hematoxylin and eosin stained on the tissue is randomly adjusted. This is done by decomposing the RGB color of the tissue into H\&E color space~\cite{ruifrok2001quantification}, followed by multiplying the magnitude of H and E of every pixel by two i.i.d. Gaussian random variables with expectation equal to one.

\SubSection{CNN architecture}

The architecture of our CNN is shown in Tab.~\ref{tab:cnn}. We used the CAFFE tool box~\cite{jia2014caffe} for the CNN implementation. The network was trained on a single NVidia Tesla 40K GPU.

\begin{table}[h!]
	\centering
	\begin{tabular}{| r | c | c |}
	\hline
	 Layer & Filter size, stride & Output W$\times$H$\times$N \\
    \hline
    \hline
	 Input & - & $400 \times 400 \times 3$ \\
	 Conv & $10 \times 10$, 2 & $196 \times 196 \times 80$ \\
	 ReLU+LRN & - & $196 \times 196 \times 80$ \\
	 Max-pool & $6 \times 6$, 4 & $49 \times 49 \times 80$ \\
	\hline
	 Conv & $5 \times 5$, 1 & $45 \times 45 \times 120$ \\
	 ReLU+LRN & - & $45 \times 45 \times 120$ \\
	 Max-pool & $3 \times 3$, 2 & $22 \times 22 \times 120$ \\
	\hline
	 Conv & $3 \times 3$, 1 & $20 \times 20 \times 160$ \\
	 ReLU & - & $20 \times 20 \times 160$ \\
	\hline
	 Conv & $3 \times 3$, 1 & $18 \times 18 \times 200$ \\
	 ReLU & - & $18 \times 18 \times 200$ \\
	 Max-pool & $3 \times 3$, 2 & $9 \times 9 \times 200$ \\
	\hline
	 FC & - & 320 \\
	 ReLu+Drop & - & 320 \\
	\hline
	 FC & - & 320 \\
	 ReLu+Drop & - & 320 \\
	\hline
	 FC & - & Dataset dependent \\
	 Softmax & - & Dataset dependent \\
	 
	\hline
	\end{tabular}
\Caption{The architecture of our CNN used in glioma and NSCLC classification. ReLU+LRN is a sequence of Rectified Linear Units (ReLU) followed by Local Response Normalization (LRN). Similarily, ReLU+Drop is a sequence of ReLU followed by dropout. The dropout probability is $0.5$.}
\label{tab:cnn}
\end{table}

\SubSection{Experiment setup}

The WSIs of 80\% of the patients are randomly selected to train the model and the remaining 20\% to test. Depending on method, training patches are further divided into i) CNN and ii) decision fusion model training sets. We separate the data twice and average the results. Tested algorithms  are:
\begin{enumerate}
\item CNN-Vote: CNN followed by voting (average-pooling). All patches extracted from a WSI are used to train the patch-level CNN. There is no second-level model. Instead, the final predicted label of a WSI is voted by the predictions of all patches.
\item CNN-SMI: CNN followed by max-pooling. Same as CNN-Vote except the final predicted label of a WSI equals to the predicted label of the patch with maximum probability over all other patches and classes.
\item CNN-Fea-SVM: We apply feature fusion instead of decision level fusion. In particular, the outputs of the second fully connected layer of the CNN on all patches are aggregated by 3-norm pooling~\cite{xu2015deep}. Then an SVM with RBF kernel is applied to predict the image-level label given fused features.
\item EM-CNN-Vote/SMI, EM-CNN-Fea-SVM: EM-based method with CNN-Vote, CNN-SMI, CNN-Fea-SVM respectively. The patch-level EM-CNN is trained on discriminative patches identified by the E-step. Depending on the dataset, the discriminative threshold $P_1$ for each image ranges from 0.18 to 0.25; the discriminative threshold $P_2$ for each class ranges from 0.05 to 0.28 (details in  Sec.~\ref{sec:our_method2}). In each M-step, the CNN is trained on all the discriminative patches for 2 epochs.
\item EM-Finetune-CNN-Vote/SMI: Similar to EM-CNN-Vote/SMI except that instead of training a CNN from scratch, we fine-tune a pretrained 16-layer CNN model~\cite{Simonyan14c} by training it on discriminative patches.
\item CNN-LR: CNN followed by logistic regression. Same as CNN-Vote except that we train a second-level multi-class logistic regression to predict the image-level label. One tenth of the patches in each image is held out from the CNN to train the second-level multi-class logistic regression.
\item CNN-SVM: CNN followed by SVM with RBF kernel instead of logistic regression.
\item EM-CNN-LR/SVM: EM-based method with CNN-LR and CNN-SVM respectively.
\item EM-CNN-LR w/o spatial smoothing: No Gaussian smoothing is applied to estimate $P(H\mid X)$. Other parts are the same as EM-CNN-LR.
\item EM-Finetune-CNN-LR/SVM: Similar to EM-CNN-LR/SVM except that instead of training a CNN from scratch, we fine-tune a pretrained 16-layer CNN model~\cite{Simonyan14c} by training it on discriminative patches.
\item SMI-CNN-SMI: CNN with max-pooling at both discriminative patch identification and image-level prediction steps. For the patch-level CNN training, in each WSI only one patch with the highest confidence is considered discriminative.
\item NM-LBP: Nuclei Morphological features~\cite{cooper2012integrated} and rotation invariant Local Binary Patterns~\cite{ojala2002multiresolution} are extracted from all patches. A Bag-of-Words (BoW)~\cite{fei2005bayesian,yang2007evaluating} feature is built using k-means followed by SVM with RBF kernel~\cite{chang2011libsvm}. This is a non-CNN baseline.
\item Pretrained-CNN-Fea-SVM: Similar to CNN-Fea-SVM. But instead of training a CNN, we use a pretrained 16-layer CNN model~\cite{Simonyan14c} to extract features from patches. Then we select the top 500 features according to accuracy on the training set~\cite{xu2015deep}.
\item Pretrained-CNN-Bow-SVM: We build a BoW model using k-means on features extracted by the pretrained CNN, followed by SVM~\cite{xu2015deep}.

\end{enumerate}

\SubSection{WSI of glioma classification}
\label{sub:glioma_results}

There are WSIs of six subtypes of glioma in the TCGA dataset~\cite{tcga}. The numbers of WSIs and patients in each class are shown in Tab.~\ref{tab:ins_num}. All classes are described in App.~\ref{app:app}.

\begin{table}[h!]
	\centering
	\begin{tabular}{| c | c | c | c | c | c | c |}
	\hline
	 Gliomas & GBM & OD & OA & DA & AA & AO \\
    \hline
     \# patients & 209 & 100 & 106 & 82 & 29 & 13 \\
     \# WSIs & 510 & 206 & 183 & 114 & 36 & 15 \\
	\hline
	\end{tabular}
\Caption{The numbers of WSIs and patients in each class from the TCGA dataset. Class descriptions are in App.~\ref{app:app}.}
\label{tab:ins_num}
\end{table}

\begin{table}[h!]
	\centering
	\begin{tabular}{| l | r | r |}
	\hline
	 Methods & Acc & mAP \\
	\hline
	\hline
	 CNN-Vote & 0.710 & 0.812 \\
	 CNN-SMI & 0.710 & 0.822 \\
	 CNN-Fea-SVM & 0.688 & 0.790 \\
	 EM-CNN-Vote & 0.733 & 0.837 \\
	 EM-CNN-SMI & 0.719 & 0.823 \\
	 EM-CNN-Fea-SVM & 0.686 & 0.790 \\
     EM-Finetune-CNN-Vote & 0.719 & 0.817 \\
     EM-Finetune-CNN-SMI & 0.638 & 0.758 \\
	\hline
	 CNN-LR & 0.752 & \textbf{0.847} \\
	 CNN-SVM & 0.697 & 0.791 \\
	 EM-CNN-LR & \textbf{0.771} & 0.845 \\
	 EM-CNN-LR w/o spatial smoothing & 0.745 & 0.832 \\
	 EM-CNN-SVM & 0.730 & 0.818 \\
     EM-Finetune-CNN-LR & 0.721 & 0.822 \\
     EM-Finetune-CNN-SVM & 0.738 & 0.828 \\
	\hline
	 SMI-CNN-SMI & 0.683 & 0.765 \\
	 NM-LBP & 0.629 & 0.734 \\
	 Pretrained CNN-Fea-SVM & 0.733 & 0.837 \\
     Pretrained-CNN-Bow-SVM & 0.667 & 0.756 \\
	\hline
     Chance & 0.513 & 0.689 \\
	\hline
	\end{tabular}
\Caption{Glioma classification results. The proposed EM-CNN-LR method achieved the best result, close to inter-observer agreement between pathologists. (Sec.~\ref{sub:glioma_results} ).}
\label{tab:brain}
\end{table}

\begin{table}[h!]
	\centering
	\begin{tabular}{r | r | r | r | r | r | r |}
	\cline{2-7}
	& \multicolumn{6}{|c|}{Predictions} \\
	\hline
	 \multicolumn{1}{|c|}{Ground Truth} & GBM & OD & OA & DA & AA & AO \\
	\hline
    \hline
     \multicolumn{1}{|c|}{GBM} & 214 & 0 & 2 & 0 & 1 & 0 \\
    \hline
     \multicolumn{1}{|c|}{OD} & 1 & 47 & 22 & 2 & 0 & 1 \\
    \hline
     \multicolumn{1}{|c|}{OA} & 1 & 18 & 40 & 8 & 3 & 1 \\
    \hline
     \multicolumn{1}{|c|}{DA} & 3 & 9 & 6 & 20 & 0 & 1 \\
    \hline
     \multicolumn{1}{|c|}{AA} & 3 & 2 & 3 & 3 & 4 & 0 \\
    \hline
     \multicolumn{1}{|c|}{AO} & 2 & 2 & 3 & 0 & 0 & 1 \\
	\hline
	\end{tabular}
\Caption{Confusion matrix of glioma classification. The nature of Oligoastrocytoma causes the most confusions. See Sec.~\ref{sub:glioma_results} for details.}
\label{tab:brainConfusion}
\end{table}

The results of our experiments are shown in Tab.~\ref{tab:brain}. The confusion matrix is given in Tab.~\ref{tab:brainConfusion}. An experiment showed that the inter-observer agreement of two experienced pathologists on a similar dataset \footnote{\label{note1}Results not directly comparable due to possible dataset differences.} was approximately 70\% and that even after reviewing the cases together, they agreed only around 80\% of the time~\cite{gupta2005clarifying}. Therefore, our accuracy of 77\% is similar to inter-observer agreement.

In the confusion matrix, we note that the classification accuracy between GBM and Low-Grade Glioma (LGG) is 97\% (chance was 51.3\%). A fully supervised method achieved 85\% accuracy using a domain specific algorithm trained on ten manually labeled patches per class~\cite{mousavi2015automated}. To the best of our knowledge our method is the first to classify five LGG subtypes automatically, a much more challenging classification task than the benchmark GBM vs. LGG classification. We achieve 57.1\% LGG-subtype classification accuracy with chance at 36.7\%. Notice that most of the confusions are related to oligoastrocytoma (OA) because it is a mixed glioma that is challenging even for pathologists to agree on, according to a neuropathology study: ``Oligoastrocytomas contain distinct regions of oligodendroglial and astrocytic differentiation... The minimal percentage of each component required for the diagnosis of a mixed glioma has been debated, resulting in poor inter-observer reproducibility for this group of neoplasms.''~\cite{brat2008diagnosis}.

We compare recognition rates for the OA subtype. The F-score of OA recognition is \textbf{0.426}, \textbf{0.482}, and \textbf{0.544} using PreCNN-Fea-SVM, CNN-LR, and EM-CNN-LR respectively. We thus see that the improvement over other methods becomes increasingly more significant using our proposed method on the harder-to-classify classes.

\begin{figure}[t]
\begin{center}
\subcaptionbox*{}{\rotatebox[origin=t]{90}{\hspace{0.2cm} GBM}}
   \begin{subfigure}[b]{0.11\textwidth}
   	\includegraphics[width=\textwidth]{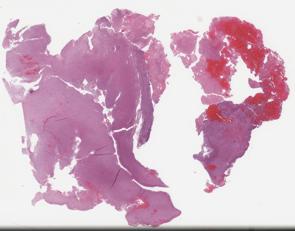}
   \end{subfigure}
   \begin{subfigure}[b]{0.11\textwidth}
   	\includegraphics[width=\textwidth]{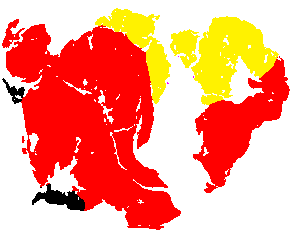}
   \end{subfigure}
   \begin{subfigure}[b]{0.11\textwidth}
   	\includegraphics[width=\textwidth]{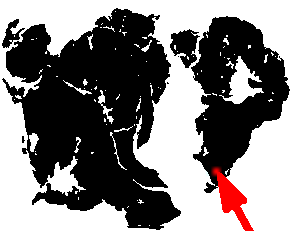}
   \end{subfigure}
   \begin{subfigure}[b]{0.11\textwidth}
   	\includegraphics[width=\textwidth]{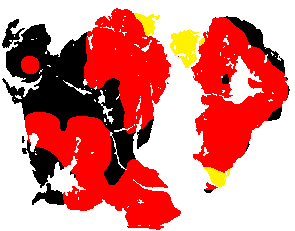}
   \end{subfigure}

\vspace{-0.5cm}
\subcaptionbox*{}{\rotatebox[origin=t]{90}{\hspace{0.5cm} GBM}}
   \begin{subfigure}[b]{0.11\textwidth}
   	\includegraphics[width=\textwidth]{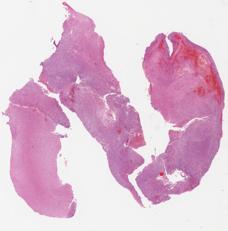}
   \end{subfigure}
   \begin{subfigure}[b]{0.11\textwidth}
   	\includegraphics[width=\textwidth]{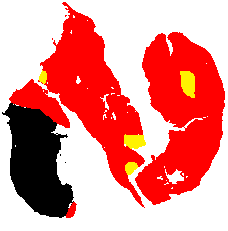}
   \end{subfigure}
   \begin{subfigure}[b]{0.11\textwidth}
   	\includegraphics[width=\textwidth]{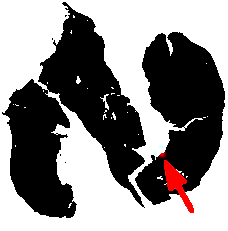}
   \end{subfigure}
   \begin{subfigure}[b]{0.11\textwidth}
   	\includegraphics[width=\textwidth]{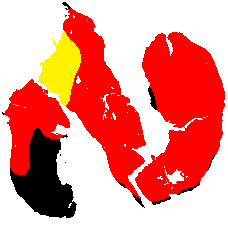}
   \end{subfigure}

\vspace{-0.5cm}
\subcaptionbox*{}{\rotatebox[origin=t]{90}{\hspace{0.5cm} OD}}
   \begin{subfigure}[b]{0.11\textwidth}
   	\includegraphics[width=\textwidth]{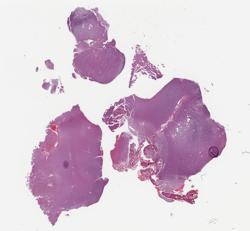}
   \end{subfigure}
   \begin{subfigure}[b]{0.11\textwidth}
   	\includegraphics[width=\textwidth]{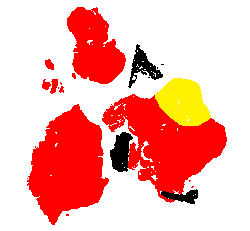}
   \end{subfigure}
   \begin{subfigure}[b]{0.11\textwidth}
   	\includegraphics[width=\textwidth]{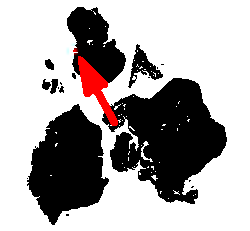}
   \end{subfigure}
   \begin{subfigure}[b]{0.11\textwidth}
   	\includegraphics[width=\textwidth]{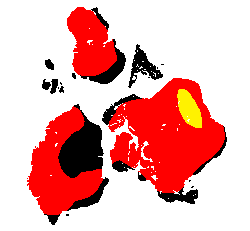}
   \end{subfigure}

\vspace{-0.5cm}
\subcaptionbox*{}{\rotatebox[origin=t]{90}{\hspace{1cm} OA}}
   \begin{subfigure}[b]{0.11\textwidth}
   	\includegraphics[width=\textwidth]{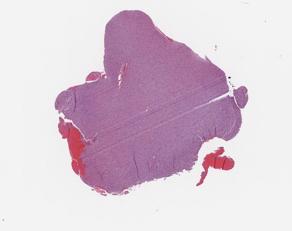}
   	\caption*{WSIs}
   \end{subfigure}
   \begin{subfigure}[b]{0.11\textwidth}
   	\includegraphics[width=\textwidth]{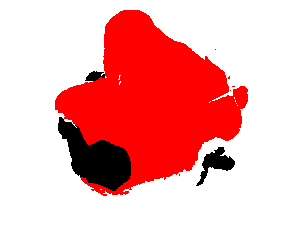}
   	\caption*{Pathologist}
   \end{subfigure}
   \begin{subfigure}[b]{0.11\textwidth}
   	\includegraphics[width=\textwidth]{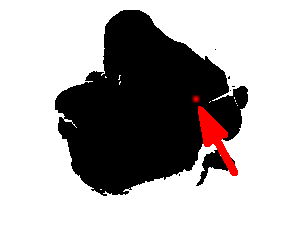}
   	\caption*{Max-pooling}
   \end{subfigure}
   \begin{subfigure}[b]{0.11\textwidth}
   	\includegraphics[width=\textwidth]{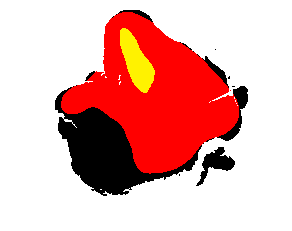}
   	\caption*{EM}
   \end{subfigure}
\end{center}
\vspace{-0.8cm}
   \Caption{Examples of discriminative patch (region) segmentation (best viewed in color). Discriminative regions are indicated in red. Diagnostic or highly discriminative regions are yellow. Non-discriminative regions are in black. Pathologist: ground truth by a pathologist. Max-pooling: results by CNN with the SMI assumption (SMI-CNN-SMI). The discriminative patches are indicated by red arrows. EM: results by our EM-based patch-level CNN (EM-CNN-Vote/SMI/LR). Notice that max-pooling does not segment enough discriminative regions.}
\label{fig:segmentation}
\end{figure}

The discriminative patch (region) segmentation results in Fig.~\ref{fig:segmentation} demonstrate the quality of our EM-based method.

\SubSection{WSI of NSCLC classification}
\label{subsec:exp_nsclc}
We use three major subtypes of Non-Small-Cell Lung Carcinoma (NSCLC). Numbers of WSIs and patients in each class are in Tab.~\ref{tab:nsclc_ins_num}. All classes are listed in App.~\ref{app:app}.
\begin{table}[h!]
	\centering
	\begin{tabular}{| c | c | c | c |}
	\hline
     NSCLCs & SCC & ADC & ADC-mix \\
    \hline
     \# patients & 347 & 291 & 80 \\
     \# WSIs & 316 & 250 & 75 \\
	\hline
	\end{tabular}
\Caption{The numbers of WSIs and patients in each class from the TCGA dataset. Class descriptions are in App.~\ref{app:app}.}
\label{tab:nsclc_ins_num}
\end{table}

\begin{table}[h!]
	\centering
	\begin{tabular}{| p{5.1cm} | r | r |}
	\hline
	 Methods & Acc & mAP \\
	\hline
	\hline
	 CNN-Vote & 0.702 & 0.838 \\
	 CNN-SMI & 0.731 & 0.852 \\
	 CNN-Fea-SVM & 0.637 & 0.793 \\
	 EM-CNN-Vote & 0.714 & 0.842 \\
	 EM-CNN-SMI & 0.731 & 0.850 \\
	 EM-CNN-Fea-SVM & 0.637 & 0.791 \\
     EM-Finetune-CNN-Vote & 0.773 & 0.877 \\
     EM-Finetune-CNN-SMI & 0.729 & 0.853 \\
	\hline
	 CNN-LR & 0.727 & 0.845 \\
	 CNN-SVM & 0.738 & 0.856 \\
	 EM-CNN-LR & 0.743 & 0.856 \\
	 EM-CNN-SVM & 0.759 & 0.869 \\
     EM-Finetune-CNN-LR & 0.784 & 0.883 \\
     EM-Finetune-CNN-SVM & \textbf{0.798} & \textbf{0.889} \\
	\hline
	 SMI-CNN-SMI & 0.531 & 0.749 \\
	 Pretrained CNN-Fea-SVM & 0.778 & 0.879 \\
     Pretrained-CNN-Bow-SVM & 0.759 & 0.871 \\
	\hline
     Chance & 0.484 & 0.715 \\
	\hline
	\end{tabular}
\Caption{NSCLC classification results. The proposed EM-CNN-SVM and EM-Finetune-CNN-SVM achieved best results, close to the inter-observer agreement between pathologists. See Sec.~\ref{subsec:exp_nsclc} for details.}
\label{tab:lung}
\end{table}

\begin{table}[h!]
	\centering
	\begin{tabular}{r | r | r | r |}
	\cline{2-4}
	& \multicolumn{3}{|c|}{Predictions} \\
	\hline
	 \multicolumn{1}{|c|}{Ground Truth} & SCC & ADC & ADC-mix  \\
	\hline
    \hline
     \multicolumn{1}{|c|}{SCC} & 199 & 26 & 0  \\
    \hline
     \multicolumn{1}{|c|}{ADC} & 30 & 155 & 11  \\
    \hline
     \multicolumn{1}{|c|}{ADC-mix} & 2 & 25 & 17  \\
    \hline
	\end{tabular}
\Caption{The confusion matrix of NSCLC classification.}
\label{tab:lungConfusion}
\end{table}

Experimental results are shown in Tab.~\ref{tab:lung}; the confusion matrix is  in Tab.~\ref{tab:lungConfusion}. When classifying SCC vs. non-SCC,  inter-observer agreement between pulmonary pathology experts and between community pathologists measured by Cohen's kappa is $\kappa=0.64$ and $\kappa=0.41$ respectively~\cite{grilley2013validation}. We achieved $\kappa=0.75$. When classifying ADC vs. non-ADC, the inter-observer agreement between experts and between community pathologists are $\kappa=0.69$ and $\kappa=0.46$ respectively~\cite{grilley2013validation}. We achieved $\kappa=0.60$. Therefore, our results appear close to inter-observer agreement \footnote{\label{note1} Results not directly comparable due to possible dataset differences.}.

The ADC-mix subtype is hard to classify because it contains visual features of multiple NSCLC subtypes. The Pretrained CNN-Fea-SVM method achieves an F-score of \textbf{0.412} recognizing ADC-mix cases, whereas our proposed method EM-Finetune-CNN-SVM achieves \textbf{0.472}. Consistent with the glioma results, our method's performance advantages are more pronounced in the hardest cases.

\SubSection{Rail surface defect severity grade classification} \label{subsec:exp_rail}

A CNN cannot be applied to gigapixel images directly because of computational limitations. We argue that even when the images are small enough for CNNs, our patch-based method compares favorably to an image-based CNN if discriminative information is encoded in image patch scale and dispersed throughout the images.

\begin{figure}
\begin{center}
   \begin{subfigure}[b]{0.236\linewidth}
   	\includegraphics[width=\textwidth]{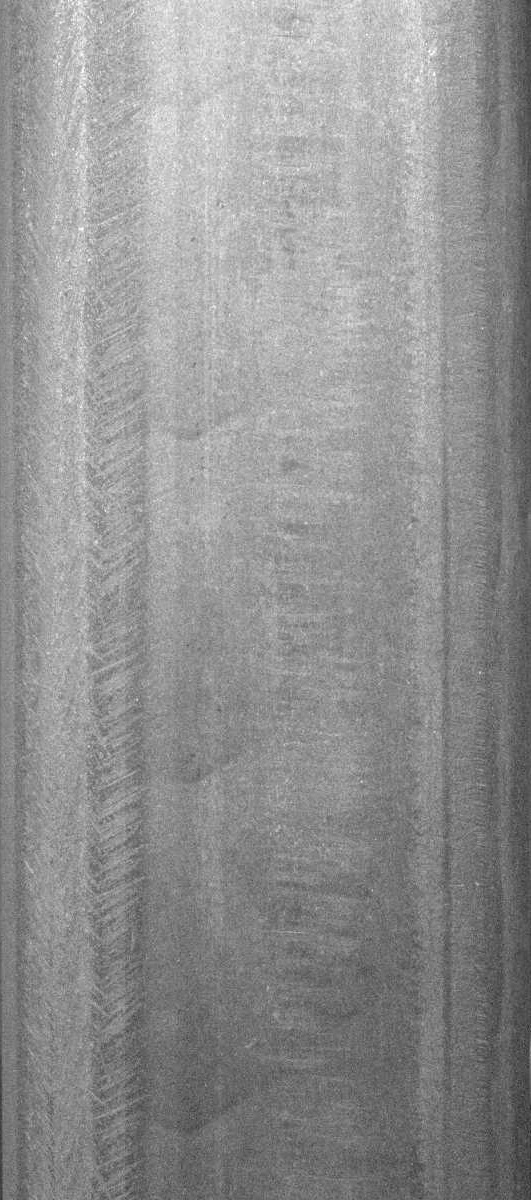}
    \vspace{-0.3cm}
   	\Caption{Grade 0}
   \end{subfigure}
   \begin{subfigure}[b]{0.2185\linewidth}
   	\includegraphics[width=\textwidth]{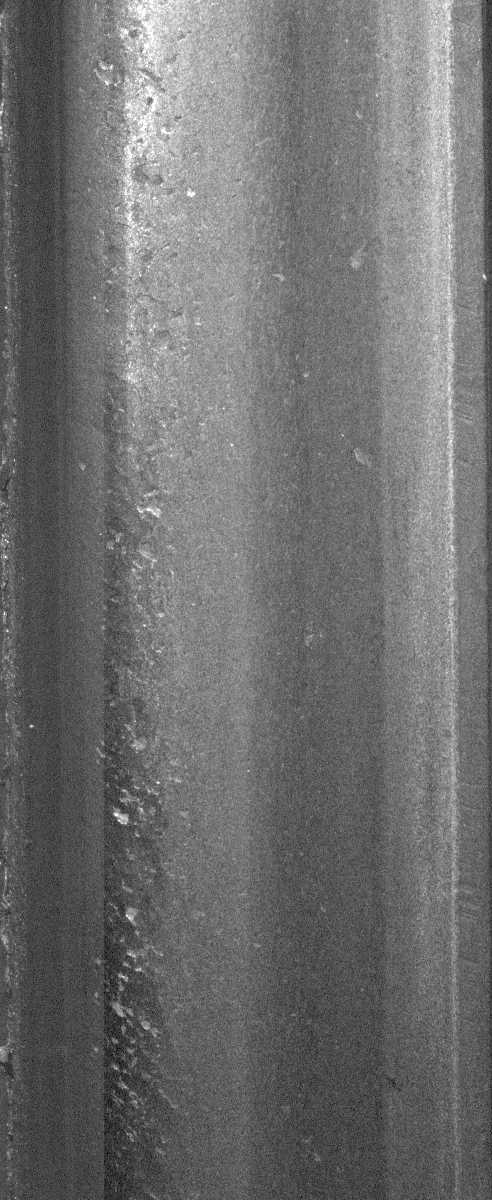}
    \vspace{-0.3cm}
   	\Caption{Grade 2}
   \end{subfigure}
   \begin{subfigure}[b]{0.2295\linewidth}
   	\includegraphics[width=\textwidth]{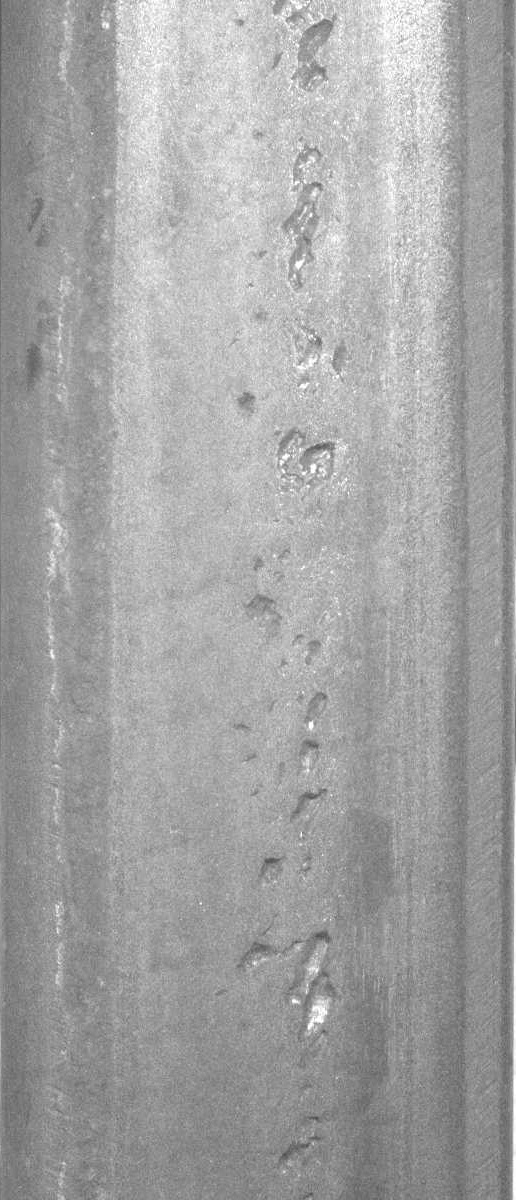}
    \vspace{-0.3cm}
   	\Caption{Grade 4}
   \end{subfigure}
   \begin{subfigure}[b]{0.245\linewidth}
   	\includegraphics[width=\textwidth]{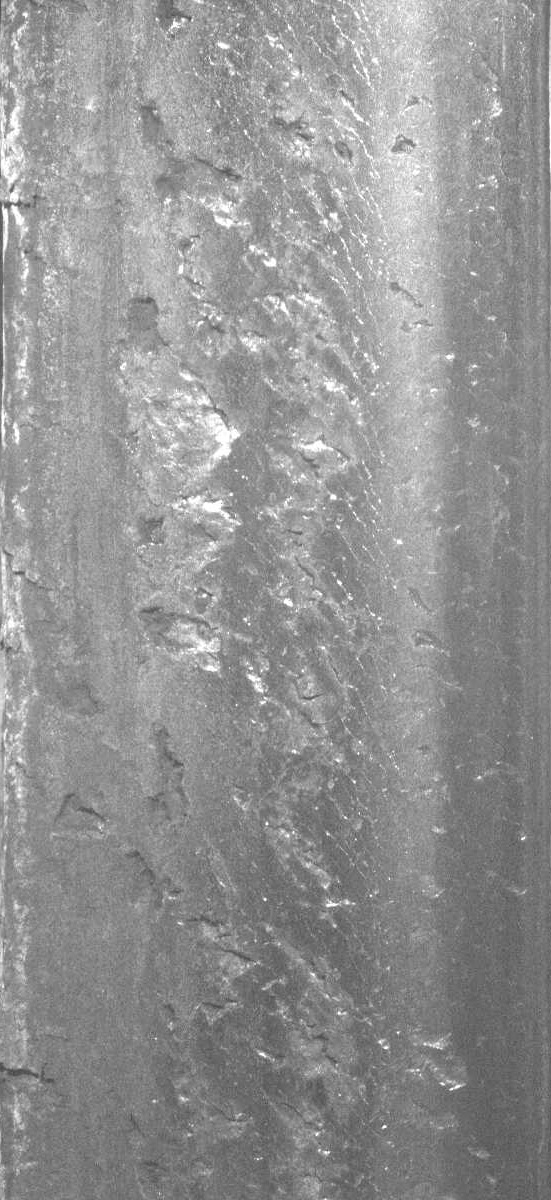}
    \vspace{-0.3cm}
   	\Caption{Grade 7}
   \end{subfigure}
\end{center}
    \vspace{-0.2cm}
   \Caption{Sample images of rail surfaces. The grade indicates defect severity. Notice that the defects are in image patch scale and dispersed throughout the image.}
\label{fig:rail_surface}
\end{figure}

To test our hypothesis, we apply our patch-based method to the task of classifying the severity grade of rail surface defects. Maintenance of rail surfaces depends on the severity of surface defects. Automatic defect grading can obviate the need for laborious  examination  and grading of rail surface defects on a regular basis. We used a dataset of 939  rail surface images with defect severity grades from 0 to 7. Typical image resolution  is 1200 by 500, as in  Fig.~\ref{fig:rail_surface}.

To support our claim, we tested two additional methods.
\begin{enumerate}
\setlength\itemsep{0.04em}
\item CNN-Image: We apply the CNN on image scale directly. In particular, the CNN is trained on 400 by 400 regions randomly extracted from images in each iteration. At test time, we apply the CNN on five regions (top left, top right, bottom left, bottom right, center) and average the predictions.
\item Pretrained CNN-ImageFea-SVM: We apply a pretrained 16-layer network~\cite{Simonyan14c} to rail surface images to extract features, and train an SVM on these features.
\end{enumerate}

The CNN used in this experiment has a similar achitecture to the one described in Tab.~\ref{tab:cnn} with smaller and fewer filters. The size of patches in our patch-based methods is 64 by 64. We apply 4-fold cross-validation and show the averaged results in Tab.~\ref{tab:rail}. Our patch-based method EM-CNN-SVM and EM-CNN-Fea-SVM outperform the conventional image-based method CNN-Image. Moreover, results using CNN features extracted on patches (Pretrained CNN-Fea-SVM) are better than results with CNN features extracted on images (Pretrained-CNN-ImageFea-SVM).

\begin{table}[h!]
	\centering
	\begin{tabular}{| p{5.1cm} | r | r |}
	\hline
	 Methods & Acc & mAP \\
	\hline
	\hline
	 CNN-Vote & 0.695 & 0.823 \\
	 CNN-SMI & 0.700 & 0.801 \\
	 CNN-Fea-SVM & 0.822 & 0.903 \\
	 EM-CNN-Vote & 0.683 & 0.817 \\
	 EM-CNN-SMI & 0.684 & 0.799 \\
	 EM-CNN-Fea-SVM & \textbf{0.830} & \textbf{0.908} \\
	\hline
	 CNN-LR & 0.764 & 0.867 \\
	 CNN-SVM & 0.803 & 0.886 \\
	 EM-CNN-LR & 0.772 & 0.871 \\
	 EM-CNN-SVM & 0.813 & 0.895 \\
	\hline
	 SMI-CNN-SMI & 0.258 & 0.461 \\
	 Pretrained CNN-Fea-SVM & 0.808 & 0.894 \\
    \hline
     CNN-Image & 0.770 & 0.876 \\
	 Pretrained CNN-ImageFea-SVM & 0.778 & 0.878 \\
	\hline
     Chance & 0.228 & 0.438 \\
	\hline
	\end{tabular}
\Caption{Rail surface defect severity grade classification results. Our patch-based method EM-CNN-SVM and EM-CNN-Fea-SVM outperform image-based methods CNN-Image and Pretrained CNN-ImageFea-SVM significantly.}
\label{tab:rail}
\end{table}

\Section{Conclusions}
\label{sec:conclusions}

We presented a patch-based Convolutional Neural Network (CNN) model with a supervised decision fusion model that is successful in Whole Slide Tissue Image (WSI) classification. We proposed an Expectation-Maximization (EM) based method that identifies discriminative patches automatically for CNN training. With our algorithm, we can classify subtypes of cancers given WSIs of patients with accuracy similar or close to inter-observer agreements between pathologists. Furthermore, we experimentally demonstrate using a comparable non-cancer dataset of smaller images, that the performance of our patch-based CNN compare favorably to that of an image-based CNN. In future work we will leverage the non-discriminative patches as part of the data likelihood in the EM formulation instead of assuming they are uniformly distributed. We will explore ways to optimize CNN-training so that it scales up to the large scale pathology datasets that are becoming available.
\begin{appendices}
\Section{Description of cancer subtypes}
\label{app:app}
The manual classification of Gliomas and Non-Small-Cell Lung Carcinomas (NSCLC) into subtypes includes assessment of cell distributions and characteristics such as shape and texture, and tissue region characteristics such as existence of necrotic regions.

\begin{description}
\setlength\itemsep{-0.2em}
\item[GBM] Glioblastoma, ICD-O 9440/3, WHO grade IV. A Whole Slide Image (WSI) is classified as GBM iff one patch can be classified as GBM with high confidence.
\item[OD] Oligodendroglioma, ICD-O 9450/3, WHO grade II.
\item[OA] Oligoastrocytoma, ICD-O 9382/3, WHO grade II; Anaplastic oligoastrocytoma, ICD-O 9382/3, WHO grade III. This mixed glioma subtype is hard to classify even by pathologists~\cite{gupta2005clarifying}.
\item[DA] Diffuse astrocytoma, ICD-O 9400/3, WHO grade II.
\item[AA] Anaplastic astrocytoma, ICD-O 9401/3, WHO grade III.
\item[AO] Anaplastic oligodendroglioma, ICD-O 9451/3, WHO grade III.
\item[LGG] Low-Grade-Glioma. Include OD, OA, DA, AA, AO.
\item[SCC] Squamous cell carcinoma, ICD-O 8070/3.
\item[ADC] Adenocarcinoma, ICD-O 8140/3.
\item[ADC-mix] ADC with mixed subtypes, ICD-O 8255/3.
\end{description}
\end{appendices}

{\small
\bibliographystyle{ieee}
\bibliography{egbib}
}

\end{document}